\documentclass[review]{elsarticlemod}
\usepackage{lineno,hyperref}
\modulolinenumbers[5]

\biboptions{authoryear}

\usepackage{amsmath}
\usepackage{algorithm} %Algorithmus
\usepackage{algorithmicx} %Algorithmus
\usepackage{algpseudocode} %Pseudocode
\usepackage{fixltx2e}
\usepackage{graphicx}
\usepackage{tabularx}
\usepackage[flushleft]{threeparttable}

\begin{document}
%\Large{
\begin{frontmatter}

\title{On the performance of different mutation operators of a subpopulation-based genetic algorithm for multi-robot task allocation problems}
%% \tnotetext[mytitlenote]{Fully documented templates are available in the elsarticle package on \href{http://www.ctan.org/tex-archive/macros/latex/contrib/elsarticle}{CTAN}.}

%%\author{Chun Liu, Andreas Kroll}
%%\address{Department of Measurement and Control, Mechanical Engineering, University of Kassel\\ M\"{o}nchebergstra{\ss}e 7, 34125, Kassel, Germany}
%% \fntext[myfootnote]{Since 1880.}

%% \author[mymainaddress,mysecondaryaddress]{Elsevier Inc}
%% \ead[url]{www.elsevier.com}

%% \author[mysecondaryaddress]{Global Customer Service\corref{mycorrespondingauthor}}
%% \cortext[mycorrespondingauthor]{Corresponding author}
%% \ead{support@elsevier.com}

%\author{Chun Liu\corref{mycorrespondingauthor}}
%\ead{chun.liu.mg@gmail.com}
%
%\author{Andreas Kroll}
%\ead{andreas.kroll@mrt.uni-kassel.de}
%
%
%\address{Department of Measurement and Control, Mechanical Engineering, University of Kassel\\ M\"{o}nchebergstra{\ss}e 7, 34125, Kassel, Germany}
\author[mymainaddress,mysecondaryaddress]{Chun Liu\corref{mycorrespondingauthor}}
%\cortext[mycorrespondingauthor]{Corresponding author. Tel.: +86 010 6228 3022}
\cortext[mycorrespondingauthor]{Corresponding author.}
\ead{chun.liu@bupt.edu.cn}

\author[mysecondaryaddress]{Andreas Kroll}
\ead{andreas.kroll@mrt.uni-kassel.de}
%mymainaddress
\address[mymainaddress]{School of Automation, Beijing University of Posts and Telecommunications\\
No 10, Xitucheng Road, 100876, Beijing, China}
\address[mysecondaryaddress]{Department of Measurement and Control, Mechanical Engineering, University of Kassel\\ M\"{o}nchebergstra{\ss}e 7, 34125, Kassel, Germany}
%% \address[mysecondaryaddress]{360 Park Avenue South, New York}

\begin{abstract}
The performance of different mutation operators is usually evaluated in conjunction with specific parameter settings of genetic algorithms and target problems. 
Most studies focus on the classical genetic algorithm with different parameters or on solving unconstrained combinatorial optimization problems such as the traveling salesman problems. 
In this paper, a subpopulation-based genetic algorithm that uses only mutation and selection is developed to solve multi-robot task allocation problems. The target problems are constrained combinatorial optimization problems, and are more complex if cooperative tasks are involved as these introduce additional spatial and temporal constraints.
The proposed genetic algorithm can obtain better solutions than classical genetic algorithms with tournament selection and partially mapped crossover. The performance of different mutation operators in solving problems without/with cooperative tasks is evaluated. 
The results imply that inversion mutation performs better than others when solving problems without cooperative tasks, and the swap-inversion combination performs better than others when solving problems with cooperative tasks. 
\end{abstract}


\end{frontmatter}

%\linenumbers

\begin{table}[h!]
\begin{footnotesize}
\begin{tabular}{p{0.15\columnwidth} p{0.78\columnwidth}}
\hline
%\multicolumn{2}{l}{Nomenclature}  \\
Symbol & Explanation \\
\hline
	$A$ & Solution of the multi-robot task allocation problem \\
	$A_k$ & Task assignment and schedule of robot $R_k$ \\
	$C_k (A_k )$ & Time for $R_k $ to complete its tasks $A_k$ \\
	$c_{ijk}^t$ & Traveling time of robot $R_k $ from inspection position of subtask $P_i $ to that of $P_j$ \\
	$c_{jk}^w$ & Waiting time of robot $R_k$ to execute subtask $P_j$ after arriving at the inspection position of $P_j$ \\
	$eli\_cnt$ & Elite count \\
	$gen\_num$ & Number of generations \\
	$J$ & Cost function (completion time) \\
	$J_{\textsubscript{max}}$ & Maximum completion time \\
	$J_{\textsubscript{mean}}$ & Mean completion time \\
	$J_{\textsubscript{min}}$ & Minimum completion time \\
	$K$ & Number of subpopulations \\
	$N^P$ & Number of subtasks \\
	$N^R$ & Number of robots \\
	$N^T$ & Number of tasks \\
	$p_a$ & Probability of producing a new gene-apportion \\
	$p_c$ & Crossover probability \\
	$p_m$ & Mutation probability \\
	$pop\_siz$ & Population size \\
	$pop\_sub$ & Subpopulation size \\
	$P$ & Set of subtasks \\
	$P_i$ & $i$-th subtask \\
	$R$ & Set of robots \\
	$R_k$ & $k$-th robot \\
	$T$ & Set of tasks \\
	$T_l$ & $l$-th task \\
	$tor\_siz$ & Tournament size \\
	$\mu$ & Mean parameter of normal distribution for producing new gene-apportions \\
	$\sigma$ & Standard deviation of normal distribution for producing new gene-apportions \\
	$\tau_i^a$ & Arrival time of robot at inspection position of $P_i$  \\
\hline
\end{tabular}
\end{footnotesize}
\end{table}

\section{Introduction}
A genetic algorithm (GA) is a centralized heuristic method inspired from biological evolution. It is widely used for optimization and search problems because of its simplicity, high flexibility in problem modeling, and good global search capability. Many genetic algorithms have been developed to solve optimization problems in bioinformatics, computational science, engineering, economics, and other fields. 
For example, in engineering applications, genetic algorithms have been used to solve the design of roof structures \citep{Kociecki2014218}, assembly problems \citep{Akpnar2011449}, and industrial plant inspection \citep{LiuKroll2012a}. 

Selection, crossover, and mutation operators maintain the population diversity \citep{McGinley2011}, and also influence the performance of genetic algorithms. Therefore, many efforts have been devoted to the design of these operators, for example,  a new selection strategy based on population recombination and elitist refinement \citep{Kwak2011}, a two-part chromosome crossover operator \citep{newx2013}, and a greedy sub tour mutation operator \citep{Albayrak2011} have been developed to improve the efficiency of genetic algorithms. Crossover and mutation are the main search operators of genetic algorithms. They play different roles in genetic algorithms: crossover tends to preserve the features of the parents, while mutation tends to make some small local perturbation of individuals. Compared to crossover, mutation is usually considered as a secondary operator with small probability in classical genetic algorithms \citep{Holland:1992}. This could be due to the fact that a large mutation rate would make genetic algorithms to search randomly. However, there has not been any theoretical proof that crossover has general advantages over mutation \citep{xvsm1992}. Many studies have shown that genetic algorithms without crossover can perform better than classical genetic algorithms, if mutation is combined with an effective selection operator \citep{Fogel90comparinggenetic, de2011, LiuKroll2012b, xvsm2014}.

Mutation is usually carried out with a single parent and plays an important role in increasing the population diversity. Various mutation operators have been developed for different solution representations, for example, Gaussian and uniform mutation for binary coding \citep{Fogel90comparinggenetic}, swap and insertion for integer coding \citep{Larranaga:1999}, polynomial and power mutation for real coding \citep{Deep2007211, realmut}. Some mutation operators are problem-dependent, such as greedy sub tour mutation for traveling salesman problems \citep{Albayrak2011} and energy mutation for multicast routing problems \citep{Karthikeyan2013}. 
Some studies suggest a mutation-combination \citep{DeepM11} or self-adaptive mutation operators \citep{dyamicm, Serpell2010, McGinley2011}.
The performance of different mutation operators has been analyzed, showing that it highly depends on the parameter choice of genetic algorithms \citep{brizuela, Wang2006, xvsm2014} and the type of problems \citep{hasan2011, Karthikeyan2013}. Most of related work has studied problems without cooperative tasks such as traveling salesman problems \citep{DeepM11, Albayrak2011} and flow shop scheduling \citep{Nearchou2004191, Wang2006}. In this paper, the performance of mutation operators will be analyzed when solving multi-robot task allocation problems with cooperative tasks.

Multi-robot task allocation (MRTA) determines the task sequence and distribution for a group of robots in multi-robot systems \citep{Gerkey2004}. It is a constrained combinatorial optimization problem, which usually provides solutions to minimize the cost while satisfying operational constraints. To find the global optimal solution, genetic algorithms \citep{LiuKroll2012a} and hybrid genetic algorithms \citep{SOCO2014} have been developed to solve MRTA problems without/with cooperative tasks. MRTA problems without cooperative tasks are similar to multiple traveling salesman problems. They are NP- (non-deterministic polynomial-time) hard optimization problems as traveling salesman problems are NP-hard. MRTA problems with cooperative tasks are more complex and strongly NP-hard \citep{Gerkey2004}, because each cooperative task requires at least two robots to carry it out simultaneously, which introduces both spatial and temporal constraints into the optimization problem.  

In this paper, a subpopulation-based genetic algorithm is developed to solve MRTA problems. This genetic algorithm deploys mutation operators and elitism selection in each subpopulation but not any crossover operator. The effects of using different mutation operators on algorithm performance are analyzed when solving MRTA problems without/with cooperative tasks.

This paper is organized as follows: multi-robot task allocation problems with cooperative tasks are introduced in Section~\ref{sec:MRTA}. Section~\ref{sec:GA} presents the subpopulation-based genetic algorithm. Simulation studies and the analysis of results are shown in Section~\ref{sec:SS}. Finally, the conclusions are drawn in Section~\ref{sec:CON}.

\section{Multi-robot task allocation problems with cooperative tasks}
\label{sec:MRTA}
Multi-robot task allocation is a combinatorial optimization problem, which assigns a set of tasks to a group of robots where typically the number of tasks is significantly larger than the number of robots. For solving this optimization problem, the first important thing is to understand what the tasks are. In general, the tasks can be classified into single-robot tasks and multi-robot tasks \citep{Gerkey2004}. A single-robot task is carried out by a single robot. A multi-robot task requires multiple robots to perform at the same time, which is also referred to as cooperative task in this paper. 
Tasks vary in different practical applications. The problem complexity increases with the number of robots required for each cooperative task.

This paper studies the problem of multi-robot task allocation for industrial plant inspection using remote sensing to detect gas and fluid leakages \citep{Bonow2013, ordonezmuller_2013c}. 
The applied sensing technology requires a diffuse reflecting background for a maximum measurement range of approximate 30 m. Larger ranges can be achieved and cases without reflecting background can be handled by using an assistant robot with a special retro-reflector \citep{ordonezmuller_2014}.
This results in two types of tasks: single- and two-robot tasks. Each single-robot task is performed by one robot with an active sensor. Each two-robot task is carried out by two robots cooperatively: the first robot with an active sensor and the second robot with a retro-reflector. Multi-robot task allocation for inspection problems with cooperative tasks introduces spatial and temporal constraints: \textit{spatial constraints}, as tasks must be executed by robots each from specific inspection position; \textit{temporal constraints}, as each cooperative task requires two robots to carry it out at the very same time.

The objective of multi-robot task allocation problems is usually to minimize the total mission cost due to energy consumption, completion time, and/or traveled distance. Inspection problems can be safety-critical, so the inspection of the whole plant is usually required to be finished as soon as possible to avoid economic loss and environmental pollution. Quicker inspection also means that the higher frequency of inspecting a plant becomes possible or that more inspection problems can be solved in a given time.
Therefore, the objective of the studied multi-robot task allocation problem in this paper is defined as the completion time. This is the time span between the first robot starting its work and the last robot finishing its tasks. 
Formally, given a set of robots $R=\{R_k \vert k\in \{1,2,...,N^R\}\}$ and a set of tasks $T=\{T_l \vert l\in \{1,2,...,N^T\}\}$, the objective (completion time) can be represented as 
\begin{equation}
\label{eq:obj}
J(A)=\mathop {\max }\limits_{k\in \{1,...,N^R\}} C_k (A_k ),
\end{equation}
where $A$ is an admissible solution of the task allocation problem, $A_k$ is the task sequence of robot $R_k$, and $C_k (A_k )$ is the time of robot $R_k$ required to finish all assigned tasks according to the sequence $A_k$.
The objective of multi-robot task allocation problems is to find the task allocation that minimizes $J(A)$. 

Denoting each single-robot task as a subtask and each cooperative task (two-robot task) as two subtasks, all subtasks would form a set $P=\{P_i \vert i\in \{1,2,...,N^P\}\}$. The task allocation $A$ must satisfy the following constraint:
\begin{equation}
\label{eq:cons}
A=\{A_k \vert \bigcup\limits_{k=1}^{N^R} {A_k } =P,A_k \bigcap\limits_{k\ne i} {A_i } =\emptyset\},
\end{equation}
with $i,k\in \{1,2,...,N^R\}$. The constraint \eqref{eq:cons} ensures that each subtask is executed only once. The task allocation must also satisfy that all robots start and end their mission at their home bases; hence, the completion time \eqref{eq:obj} also involves the traveling time of each robot from its home base to its first task and the traveling time of each robot for returning from its last task to its home base.

In addition, the following three executability constraints (EC) must also be satisfied to ensure that the task allocation is feasible for execution:
\begin{enumerate}[(EC1)]
\item Each cooperative task is carried out by two different robots.
\item Two subtasks of each cooperative task are started at the same time.
\item The schedule of cooperative tasks is feasible for execution, i.e., the sequence of cooperative tasks is not contradictory.
\end{enumerate}
For instance, two cooperative tasks $T_1$ and $T_2$ are assigned to robots $R_k$ and $R_s$. Robot $R_s$ must carry out $T_1$ first if $R_k$ perform $T_1$ first; otherwise, the solution is infeasible.

Based on the described characteristics and definition of multi-robot task allocation for inspection problems, it is obvious that the completion time includes the traveling time between tasks, the inspection time of each task, and the waiting time occurring when performing cooperative tasks. In this work, the traveling time is calculated using the A* algorithm for a given inspection environment. The inspection time is predefined according to the inspection method and measurement system properties.
The waiting time depends on the solution itself and is calculated for each solution candidate during the execution of the genetic algorithm.

\section{Subpopulation-based genetic algorithm}
\label{sec:GA}
A subpopulation-based genetic algorithm is developed to solve the multi-robot task allocation problems with cooperative tasks in this section. At the beginning of this section, the solution representation is introduced. After that, the implementation of the proposed genetic algorithm is illustrated. At the end, the subpopulation-based and a classical genetic algorithm \citep{Taplin2005} are compared.
% Page 118-119

\subsection{Solution representation}
Permutation coding is used to represent a solution of this optimization problem, because it is the most natural and readable way to represent a task sequence. This representation can be very easily implemented in the most commonly used programming languages such as MATLAB or C/C\texttt{++}. Using permutation coding, a solution of multi-robot task allocation problems with cooperative tasks is composed of $N^R$ task sequences that involve the distribution and schedule of all subtasks for all robots, which satisfy the constraint \eqref{eq:cons} in Section~\ref{sec:MRTA}. Based on the permutation coding, four coding strategies have been developed in a previous study \citep{SOCO2014} that focussed on the development and comparative analysis of these coding strategies. In this paper, one of these coding strategies, the task-based coding, will be selected to evaluate the effects of different mutation operators when solving multi-robot task allocation with cooperative tasks. This coding strategy was selected as it does not create infeasible solutions and as a genotype corresponds to just one phenotype. This avoids that the impact of the decoding strategy dilutes the performance comparison regarding mutation operators. In the following part, the encoding and the decoding of the task-based coding strategy are illustrated, respectively.

\begin{figure}[h!]
\centering
\includegraphics[width=0.4\columnwidth]{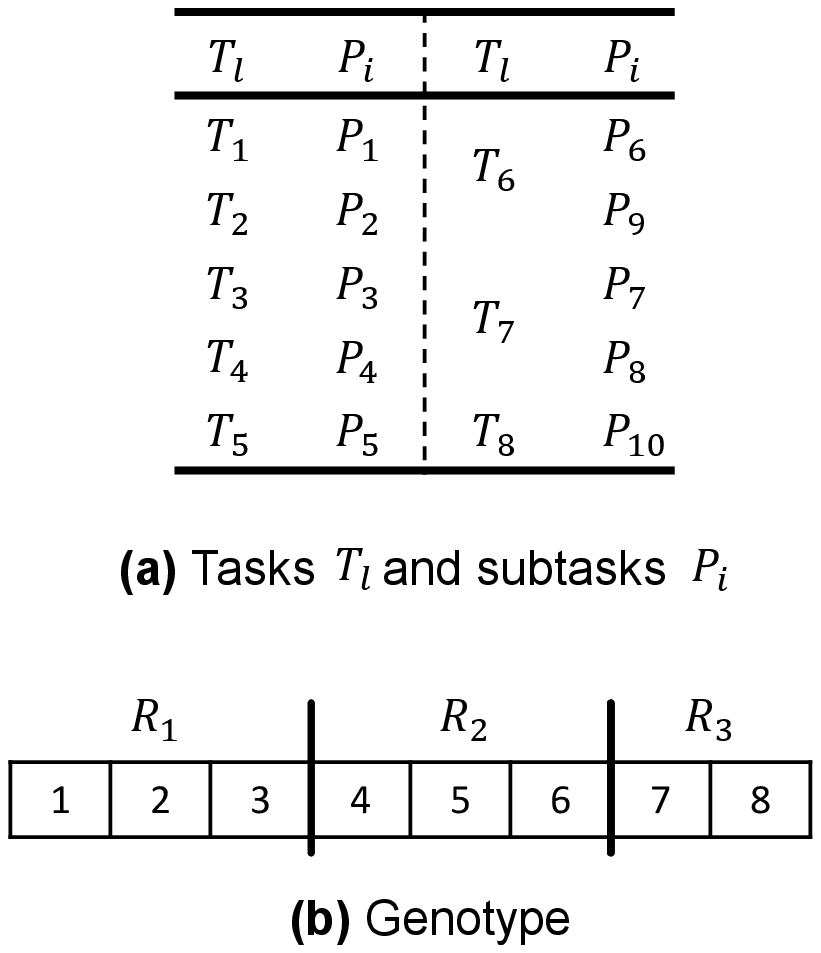}
\caption{An example with single-robot tasks ($T_1-T_5,T_8$) and cooperative tasks ($T_6,T_7$)}
\label{fig:geno}
\end{figure}

\paragraph{Encoding} 
Each gene represents a task. The genotype of a solution is composed of two parts:
\begin{itemize}
\item A \textit{chromosome} is a string of genes and represents the sequence of all $N^T$ tasks.
\item A \textit{gene-apportion} is a set of $N^R-1$ integers, which splits a chromosome into $N^R$ parts for $N^R$ robots.
\end{itemize}
For example, Fig.~\ref{fig:geno}(a) shows an example problem where six single-robot tasks and two cooperative tasks will be carried out by three robots. 
As each task is encoded as one gene, the genotype of a solution can be represented as shown in Fig.~\ref{fig:geno}(b): the chromosome $\{1,2,3,4,5,6,7,8\}$ is split into three segments by a gene-apportion $\{3,6\}$ that is represented as two vertical lines.

\paragraph{Decoding} 
A genotype is decoded as a phenotype via two steps:
\begin{enumerate}[(1)]
\item For single-robot tasks, each gene is directly decoded as its corresponding task; see Fig.~\ref{fig:lwde}(a).
\item For cooperative tasks, two subtasks of each cooperative task should be decoded. According to the genotype, it is obvious that each cooperative task is already assigned to a robot $R_k$, e.g. $T_6 $ is assigned to $R_k =R_2$. Hence, the next step is to find the second robot so that two robots can carry it out cooperatively (satisfying the constraint EC1). For each cooperative task, the decoding is: 
\begin{enumerate}[(S1)]
\item The closest subtask is assigned to robot $R_k$ based on the traveling time $c_{ijk}^t$ of robot $R_k $ from one subtask $P_i $ to another subtask $P_j$. 
\item The other subtask is inserted at the ``best'' position of the task sequences of robots except $R_k$. The ``best'' position is the position that provides the least waiting time for performing this cooperative task, which is calculated by enumerating all possible positions of the task sequences of robots except $R_k$. 
This decoding is carried out starting from the cooperative task that a robot meets first, so that all decoded phenotypes are feasible for execution.
\end{enumerate}
%This decoding is named as least-waiting-time decoding. 
\begin{figure}[h!]
\centering
\includegraphics[width=\columnwidth]{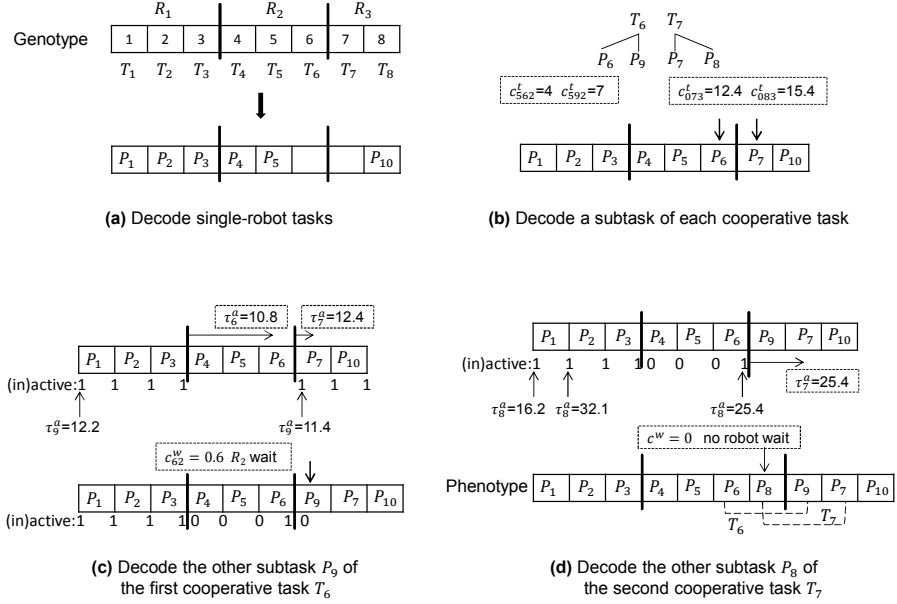}
\caption{Decoding for a genotype in Fig.~\ref{fig:geno}}
\label{fig:lwde}
\end{figure}
For instance, the decoding outcome of the step ``S1'' is shown in Fig.~\ref{fig:lwde}(b): $P_6 $ is assigned to $R_2$ because $c_{562}^t<c_{592}^t$; $P_7 $ is assigned to $R_3$ after leaving its home base (denoted as ``$0$'') because $c_{073}^t<c_{083}^t$. The decoding outcome of the step ``S2'' is shown in Fig.~\ref{fig:lwde}(c) and (d). Possible positions are marked as ``active'' (denoted as ``1''), whereas impossible positions are marked as ``inactive'' (denoted as ``0''). 
The decoding algorithm first finds the ``best'' position for $ P_9 $ because robot $ R_2 $ meets task $ T_6$ earlier than robot $ R_3 $ meets task $ T_7$, i.e., $\tau_6 ^a < \tau_7 ^a$.
There are seven possible positions for decoding $P_9$ except positions of robot $R_2$, but only five positions will be tested: for $R_1$, all four positions are tested; for $R_3$, only the position before $P_7$ is tested because $P_7$ belongs to another cooperative task, which is performed in order to satisfy the constraint EC3. As $P_9 $ being inserted before $P_7$ provides the minimum waiting time, $R_2$ is waiting for $c_{62}^w = \tau _9^a - \tau _6^a = 0.6$ at $P_6$ until $R_3$ arrives at $P_9$ such that robots $R_2$ and $R_3$ can cooperatively perform $T_6$ (satisfying the constraints EC1 and EC2). 
In order to satisfy the constraint EC3, positions of chromosomes before either $P_6$ or before $P_9$ are marked as ``inactive''; see Fig.~\ref{fig:lwde}(c). Therefore, only five active positions can be tested when decoding the next cooperative task $P_8$ (see Fig.~\ref{fig:lwde}(d)).
Before assigning $P_8$, the arriving time of $R_3$ at $P_7$ is recalculated, $\tau _7^a = 25.4$.
The minimum waiting time $c^w=0$ can be obtained when $P_8 $ is inserted after $P_6$. That is, $R_2$ arrives at $P_8$ and $R_3$ arrives at $P_7$ at the same time.
The complete task allocation obtained using this decoding requires robots $R_2$ and $R_3$ as a coalition to execute $T_6$ and $T_7$ as shown in Fig.~\ref{fig:lwde}(d).
\end{enumerate}
As illustrated above, this decoding can satisfy all executability constraints, i.e., all decoded phenotypes are feasible for execution.

Using this representation, each individual (solution candidate) includes a genotype and a phenotype.
In the proposed genetic algorithm, the chromosomes of the genotypes are mutated for generating offspring; phenotypes are used to calculate the fitness.

\subsection{Process of the developed genetic algorithm}
The developed genetic algorithm in this paper is based on subpopulations. The main idea of the genetic algorithm is that selection and mutation are applied separately in each subpopulation. 
This genetic algorithm performed well when solving medium-scale traveling salesman problems \citep{LiuKroll2012b}. In this paper, the subpopulation-based genetic algorithm is used to solve multi-robot task allocation problems.
The pseudo code of our proposed subpopulation-based genetic algorithm is presented in Algorithm~\ref{alg:ga}.

%\begin{figure} 
%\centering
\begin{algorithm}
\caption{Subpopulation-based genetic algorithm}
\label{alg:ga}
\begin{small}
  \begin{algorithmic}[1]
    \State Set parameters and select mutation operators
    \State Generate an initial population
    \While{termination criterion is not satisfied}
        \For{each genotype}
        	\State Decode it and calculate the fitness value of its phenotype
        \EndFor
        \State Divide the population into $K$ non-overlapping subpopulations randomly
        \For{each subpopulation}
        	\State Pass the $eli\_cnt$ superior individuals to the next generation directly
            \State Select the $best\_num$ superior individuals as parents
            \For{each parent}
            \State $n \leftarrow 0$
            \Repeat
            	\State $n \leftarrow n+1$
            	\State Apply mutation operator to its chromosome with a probability $p_m$
            	\State Generate the new gene-apportion with a probability $p_a$
            \Until{$n=(pop\_sub-eli\_cnt)/best\_num$}
            \EndFor
        \EndFor
        \State Form the new population from all offspring in all subpopulations
    \EndWhile
  \end{algorithmic}
\end{small}
\end{algorithm}
%\end{figure} 

\paragraph{Parameters} Parameters of the genetic algorithm are set at the beginning, such as population size ($pop\_siz$), subpopulation size ($pop\_sub$), elite count ($eli\_cnt$), mutation probability ($p_m$), and termination criterion ($gen\_num$). 

\paragraph{Initial population} The initial population is randomly produced based on the permutation coding, that is, both the chromosome and gene-apportion of each genotype in the initial population is generated at random.

\paragraph{Fitness calculation} All genotypes should be decoded as phenotypes according to the decoding procedure before fitness calculation. The fitness value of each individual is calculated according to the objective function \eqref{eq:obj}.

\paragraph{New population} As can be seen from Algorithm~\ref{alg:ga}, a new population is generated based on subpopulations. First, the whole population is randomly divided into non-overlapping subpopulations, and each subpopulation involves $pop\_sub$ individuals. After that, the elitism selection and mutation operators are applied to each subpopulation. 
The $eli\_cnt$ superior individuals are transferred to the new population, and the $best\_num$ superior individuals are selected as parents.
The $pop\_sub-eli\_cnt$ offspring are produced by mutating parents and generating new gene-apportions:
\begin{itemize}
\item The chromosome of a new offspring is produced by swap, insertion, inversion, or displacement mutation operators. 
\textit{Swap} mutation exchanges two randomly selected genes. 
\textit{Insertion} mutation moves a randomly chosen gene to another randomly chosen place. 
\textit{Inversion} mutation reverses a randomly selected gene string.
\textit{Displacement} mutation inserts a random string of genes in another random place. Insertion can be considered as a special displacement. 
\item The gene-apportion of a new offspring is generated with a probability $p_a$; otherwise, the gene-apportion of the parent is kept for the offspring. A gene-apportion is defined by $N^R-1$ integers. Each element in a new gene-apportion is generated by rounding a number that is randomly selected within the range of $[1,N^T]$ according to a standard normal distribution ($\mu, \sigma ^2$). $\mu$ is the cumulative average of the gene-apportion of the best individual obtained in each previous generation; $\sigma =0.03N^T$ is used in this paper. This gene-apportion procedure will choose numbers, which are near to the cumulative average, with a higher probability.
\end{itemize}

\paragraph{Termination criterion} The genetic algorithm is terminated when the number of generations reaches a predefined number of generations ($gen\_num$) in this paper. Both the population size and the number of generations are fixed in the simulation studies, i.e., the number of all produced individuals is constant.
There are many alternative choices of the termination criterion, e.g. maximal number of generations, CPU time limit, and fitness limit/stall. In this paper, a fixed number of generations is used because (1) CPU time highly depends on the computer hardware, (2) what is a good fitness value is unpredictable, and (3) the convergence properties are uncertain.

\subsection{Comparison of the subpopulation-based genetic algorithm and classical genetic algorithms}
The main difference between the subpopulation-based genetic algorithm and classical genetic algorithms \citep{Mitchell, Whitley2001817} is the way of producing offspring. 
The parent selection of the proposed genetic algorithm can be considered as an extended tournament selection: superior individuals in each subpopulation are selected as parents. Hence, we compare the subpopulation-based genetic algorithm with classical genetic algorithms with tournament selection (see Algorithm~\ref{alg:gat}).
The number of parents in the proposed genetic algorithm is $best\_num \cdot pop\_siz/pop\_sub$, while more parents $pop\_siz-eli\_cnt$ should be selected in classical genetic algorithms. Elites of classical genetic algorithms are chosen according to the fitness of all individuals in the whole population, while elites of the subpopulation-based genetic algorithm are selected according to the fitness of individuals in a subpopulation. Classical genetic algorithms can only keep $eli\_cnt$ best individuals, while the subpopulation-based genetic algorithm may keep local optima that may increase the population diversity. 

\begin{algorithm}
\begin{small}
\caption{A classical genetic algorithm with tournament selection}
\label{alg:gat}
  \begin{algorithmic}[1]
    \State Generate an initial population
    \While{termination criterion is not satisfied}
        \For{each genotype}
        	\State Decode it and calculate the fitness value of its phenotype
        \EndFor
        \State Pass the $eli\_cnt$ superior individuals to the next generation directly
        \Repeat
           	\State Select $tor\_siz$ individuals randomly, the best of which is chosen as a parent 
        \Until{$pop\_sub-eli\_cnt$ times}
        \State Apply crossover to the chromosomes of each pair of parents with a probability $p_c$
        \State Apply mutation to the chromosome of each offspring obtained by crossover with a probability $p_m$
        \State Generate the new gene-apportion for each offspring with a probability $p_a$
        \State Form the new population (all offspring)
    \EndWhile
  \end{algorithmic}
\end{small}
\end{algorithm}

Both crossover and mutation are used to produce offspring in classical genetic algorithms: crossover is applied with a high probability and mutation is applied with a small probability \citep{Larranaga:1999, CIAKroll2013}. Crossover dominates the search progress of classical genetic algorithms, which could be due to the fact that mutation trends to a random search. In the proposed genetic algorithm, only mutation operators are performed with a probability of $p_m=1$, which could be effective when combined with the proposed selection strategy. 
The $best\_num$ superior individuals in each subpopulation are mutated, while the rest is not used to produce offspring. The genetic algorithm exploits the solution space near to these superior individuals in this way. Non-overlapping subpopulations maintain local optima that keep diversity of the population and avoid premature convergence caused by a single superior individual. The performance of the subpopulation-based genetic algorithm will be analyzed in the next section.

The procedure of generating a new population in classical genetic algorithms is more complex than that in the proposed subpopulation-based genetic algorithm.
The time complexity of the selection in classical genetic algorithms is $O(pop\_siz-eli\_cnt)$, because $pop\_siz-eli\_cnt$ parents are selected. As illustrated above, the time complexity of the selection in the subpopulation-based genetic algorithm is $O(best\_num \cdot pop\_siz/pop\_sub)$. The time complexity of swap is O(1) as it is independent of the chromosome length. The time complexity of insertion, inversion, and displacement is $O(N^T)$ as in the worst case all genes have to be changed. 
Crossover is more complex than the above four mutation operators. Taking partially mapped crossover (PMX) \citep{Larranaga:1999} as an example, the mapping relationship between selected $numg$ genes from each pair of parents should be built to legalize the offspring. The time complexity of PMX is $O(numg+N^T)$ in the worst case: all $numg$ genes should be mapped from one parent to the other and all genes have to be changed. As crossover is applied with a higher probability, classical genetic algorithms require more CPU time than the subpopulation-based genetic algorithm.

\section{Simulation studies and analysis}
\label{sec:SS}
\begin{figure}
\centering
\includegraphics[width=0.53\columnwidth]{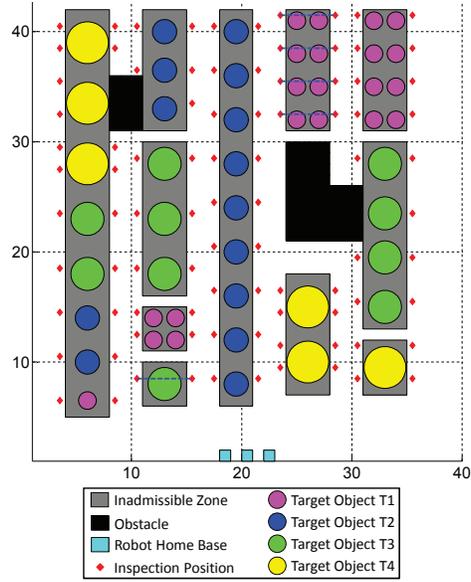}
\caption{Inspection area and tasks of Prob.C (two subtasks of each cooperative task linked by a dashed line)}
\label{fig:trow}
\end{figure}

\begin{figure}
\centering
\includegraphics[width=0.53\columnwidth]{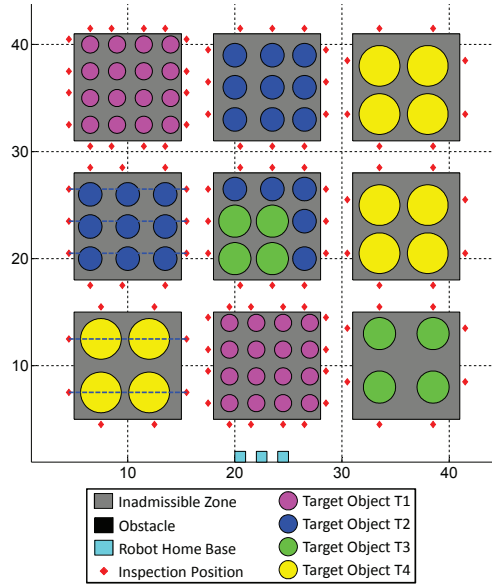}
\caption{Inspection area and tasks of Prob.D (two subtasks of each cooperative task linked by a dashed line)}
\label{fig:tisl}
\end{figure}

In this section, the performance of the proposed genetic algorithm is analyzed when solving multi-robot task allocation problems without/with cooperative tasks. Four problems are tested in the simulation studies:
\begin{itemize}
\item \textit{Prob.A} involves $90$ single-robot tasks that are distributed in rows; its inspection area is similar to that shown in Fig.~\ref{fig:trow} but all tasks are single-robot tasks.
\item \textit{Prob.B} involves $100$ single-robot tasks that are distributed in islands; its inspection area is similar to that shown in Fig.~\ref{fig:tisl} but all tasks are single-robot tasks.
\item \textit{Prob.C} involves $80$ single-robot tasks and $5$ cooperative tasks, and all tasks are distributed in rows; see Fig.~\ref{fig:trow}.
\item \textit{Prob.D} involves $90$ single-robot tasks and $5$ cooperative tasks, and all tasks are distributed in islands; see Fig.~\ref{fig:tisl}.
\end{itemize}

These scenarios have been used as test cases already in \citep{SOCO2014, phdliu} to compare the performance of different encoding and decoding strategies.
Prob.A and Prob.B are multi-robot task allocation problems without cooperative tasks. Prob.C and Prob.D are multi-robot task allocation problems with cooperative tasks. 

In the experiments, each tested genetic algorithm is performed with a population size of $pop\_siz=200$ and the number of generations chosen as $gen\_num=10^4$. 
To statistically evaluate the performance of the proposed genetic algorithm, $20$ independent runs of each algorithm are implemented on an Intel Core i3 PC with 3.2 GHz, 8 GB (RAM), Windows 7 Professional, MATLAB R2011b. More runs could provide more accurate results but require more CPU time. Hence, $20$ independent runs are carried out to restrict the computational effort, and analysis of variance (ANOVA) is used to check whether the performance differences (solution quality) between the different genetic algorithms are statistically significant.
If the value of the significance level is smaller than $0.05$, the effects of genetic algorithms are assessed to be statistically significant at a level of confidence of $95\%$.

\subsection{Case study 1: Subpopulation-based vs. binary tournament GA}
The first case study compares the performance of the subpopulation-based genetic algorithm with a classical genetic algorithm with binary tournament selection. The frameworks of both genetic algorithms are displayed in Algorithm~\ref{alg:ga} and Algorithm~\ref{alg:gat} in Section~\ref{sec:GA}, and the parameters of two genetic algorithms are listed in Table~\ref{table:para}. Inversion mutation is used in both genetic algorithms because it performs better than other mutation operators when solving combinatorial optimization problems without cooperative tasks \citep{ Wang2006, Albayrak2011, DeepM11, LiuKroll2012b}.
\begin{table}[h!]
\centering
\caption{Parameter choice in the experiments}
\label{table:para}
\begin{small}
\begin{tabular}{lll}
\hline
Parameter& Subpopulation-based GA & Classical GA\\
\hline
$pop\_sub$&  10& -- \\
$tor\_siz$&  --& 2 \\
$eli\_cnt$&  2& 2 \\
$best\_num$&  1& -- \\
$p_c$&  --& 0.9 \\
$p_m$&  1 & 0.01  \\
$p_a$ & 0.2 & 0.2 \\
Crossover & -- & PMX \\
Mutation & Inversion & Inversion \\
\hline
\end{tabular}
\end{small}
\end{table}

\begin{table}[h!]
\centering
\scriptsize{
\caption{Completion time $J$ in sec. and average CPU time in sec. for different genetic algorithms}
\label{table:case1}
\begin{threeparttable}
\begin{tabularx}{0.8\textwidth}{p{1cm}p{1cm}ll}
\hline
Problem & Criterion & Subpopulation-based GA & Classical GA \\
\hline
Prob.A& $J_{\textsubscript{min}}$& 170.06&	250.03\\
& $J_{\textsubscript{mean}}$ & 189.55&	290.07\\
& $J_{\textsubscript{max}}$& 225.56&	319.12\\
& CPU & 988	&1432\\
\hline
Prob.B& $J_{\textsubscript{min}}$& 185.95&	257.16\\
& $J_{\textsubscript{mean}}$& 207.03&	300.45\\
& $J_{\textsubscript{max}}$& 228.75&	355.11\\
& CPU & 1028&	1423\\
\hline
Prob.C&$J_{\textsubscript{min}}$ & 252.72&	348.52\\
&$J_{\textsubscript{mean}}$ & 292.78&	414.79\\
& $J_{\textsubscript{max}}$& 376.46&	500.94\\
& CPU & 2419&	2732\\
\hline
Prob.D& $J_{\textsubscript{min}}$& 255.96&	374.93\\
& $J_{\textsubscript{mean}}$& 333.07&	448.25\\
& $J_{\textsubscript{max}}$&383.95&	480.42\\
& CPU & 2580&	2885\\
\hline 
\end{tabularx}
\begin{tablenotes}
%\item [1] Here is the first Note, the width
\item
$J_{\textsubscript{max}}$ -- Maximum completion time;
$J_{\textsubscript{mean}}$ -- Mean completion time;
$J_{\textsubscript{min}}$ -- Minimum completion time.
\end{tablenotes}
\end{threeparttable}}
\end{table}

\begin{figure}[h!]
\centering
\includegraphics[width=0.95\columnwidth]{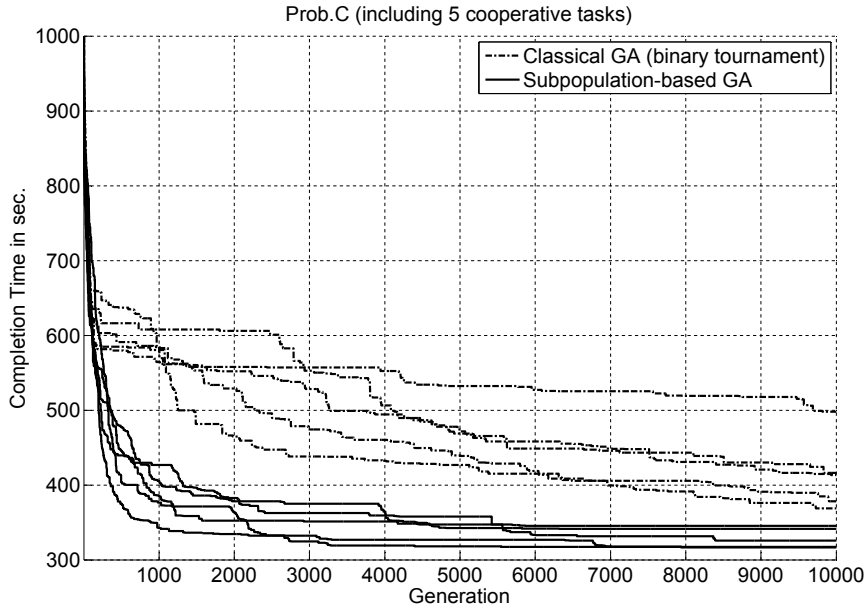}
\caption{The search progress of two genetic algorithms for solving Prob.C (5 runs selected from the total 20 runs of each algorithm)}
\label{fig:1}
\end{figure}

The experimental results are recorded in Table~\ref{table:case1}, which indicate that the proposed subpopulation-based genetic algorithm provides better solutions and requires less CPU time than the classical genetic algorithm. An ANOVA test shows that the differences of the solution quality between these two genetic algorithms are statistically significant. 
Randomly choosing 5 from the 20 runs of each genetic algorithm, the solution quality (completion time) of the best solution candidate in each generation is shown in Fig.~\ref{fig:1}. It is obvious that the subpopulation-based genetic algorithm converges significantly faster than the classical genetic algorithm within the first $1000$ generations.

\begin{figure}[h!]
\centering
\includegraphics[width=\columnwidth]{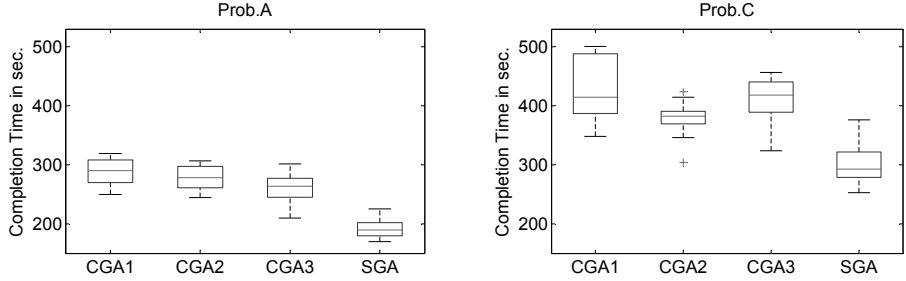}
\caption{The distribution of the solution quality of different genetic algorithms for solving Prob.A and Prob.C (20 runs): SGA -- subpopulation-based GA in Table~\ref{table:para}; 
CGA1/CGA2/CGA3 -- classical GA in Table~\ref{table:para} but GA2 with $tor\_siz=10$, GA3 with $p_c=0$ and $p_m=1$}
\label{fig:clavssub}
\end{figure}

This case study indicates that the proposed genetic algorithm based on subpopulations perform better than the classical genetic algorithm with PMX crossover and tournament selection when solving multi-robot task allocation problems, especially when requiring less CPU time and less generations.
The subpopulation-based genetic algorithm can also provide significantly better solutions than the following two classical genetic algorithms: 
(1) a classical genetic algorithm with a large tournament size $tor\_siz=10$;
(2) a classical genetic algorithm without PMX crossover and with only inversion mutation ($p_m=1$). 
The experimental results of solving Prob.A and Prob.C are shown in Fig.~\ref{fig:clavssub}, indicating that the genetic algorithm only employing mutation must be combined with an effective selection, so that the algorithm can perform well.

As discussed in Section~\ref{sec:GA}, the proposed genetic algorithm selects parents based on subpopulations that performs well in conjunction with mutation operators.
Performing the selection of the elites in each subpopulation can enhance the exploration in the search space, because it keeps both the global and local optimal solutions that avoid the algorithm being dominated by a single superior individual.  
Only the best individual in each subpopulation is chosen as the parent of mutation. In this way, the search space near the parents can be well exploited.

We also test the subpopulation-based genetic algorithm with different probabilities for generating new gene-apportions $p_a=\{0,0.2,0.4,0.6,0.8,1\}$, and the results in Table~\ref{table:probpa} show that the differences of the solution quality are not statistically significant when solving all investigated problems. It is obvious that a large $p_a$ results in more CPU time. Therefore, $p_a=0.2$ is chosen in the following experiments for analyzing the mutation effects of the subpopulation-based genetic algorithm.
The subpopulation-based genetic algorithm with a different number of parents $best\_num=\{1,2,4\}$ is performed to solve all investigated problems; see Table~\ref{table:bstn}. The results of solving Prob.A and Prob.B show that the proposed algorithm with $best\_num=\{1,2\}$ performs better than with $best\_num=4$. However, the differences between $best\_num=\{1,2,4\}$ are not statistically significant when solving Prob.C and Prob.D. In the following case studies, $best\_num=1$ is chosen.

\begin{table}[h!]
\centering
\scriptsize{
\caption{Average completion time $J_{\textsubscript{mean}}$ in sec. for the subpopulation-based genetic algorithm with different $p_a$}
\label{table:probpa}
\begin{threeparttable}
\begin{tabularx}{0.9\textwidth}{llllllp{1.5cm}l}
\hline
$p_a$ & 0 & 0.2 & 0.4 & 0.6 & 0.8 & 1 & Sig.Level*\\
\hline
Prob.A& 192.78&	189.55&	196.62&	200.20&	200.89&	196.03&	0.25\\
Prob.B& 207.98&	207.03&	205.02&	204.52&	209.12&	212.85&	0.25\\
Prob.C&294.18&	292.78&	299.76&	298.32&	305.86&	290.78&	0.95\\
Prob.D& 299.03&	308.42&	306.71&	300.72&	301.82&	295.35&	0.63\\
\hline 
\end{tabularx}
\begin{tablenotes}
\item
* The significance level obtained by an ANOVA test. If the value of Sig.Level is smaller
than $0.05$, the differences of solution quality between these $p_a$ are statistically significant.
\end{tablenotes}
\end{threeparttable}}
\end{table}

\begin{table}[h!]
\centering
\scriptsize{
\caption{Average completion time $J_{\textsubscript{mean}}$ in sec. for the subpopulation-based genetic algorithm with different $best\_num$}
\label{table:bstn}
\begin{threeparttable}
\begin{tabularx}{0.8\textwidth}{p{1.5cm}p{1.5cm}p{1.5cm}p{1.5cm}l}
\hline
$best\_num$ & 1 & 2 & 4 & Sig.Level*\\
\hline
Prob.A& 189.55 & 188.92 & 204.96 & 0.00\\
Prob.B&207.03&210.83&216.94& 0.02\\
Prob.C&292.78& 310.35& 305.20&0.76\\
Prob.D&  333.07 &340.78 & 328.4872 & 0.54 \\
\hline 
\end{tabularx}
\begin{tablenotes}
\item
* The significance level obtained by an ANOVA test. If the value of Sig.Level is smaller
than $0.05$, the differences of solution quality between these $best\_num$ are statistically significant.
\end{tablenotes}
\end{threeparttable}}
\end{table}

\subsection{Case study 2: Subpopulation-based GA with single mutation operator}
The second and the third case studies analyze the effects of the subpopulation-based genetic algorithm with different mutation operators and their combinations. Swap, insertion, inversion, and displacement mutation operators are investigated in this paper. The tested subpopulation-based genetic algorithms are listed in Table~\ref{table:case2}.
\begin{table}[h!]
\centering
\caption{Subpopulation-based genetic algorithm with different mutation operators}
\label{table:case2}
\begin{small}
\begin{tabular}{ll}
\hline
Genetic algorithm& Mutation operator(s) \\
\hline
GA1&  Swap\\
GA2&  Insertion \\
GA3&  Inversion\\
GA4&  Displacement\\\hline
GA5&  Swap and inversion\\
GA6&  Insertion and inversion \\
GA7 & Displacement and inversion \\
GA8 & Swap, insertion, inversion, and displacement\\
\hline
\end{tabular}
\end{small}
\end{table}

\begin{figure}[h!]
\centering
\includegraphics[width=\columnwidth]{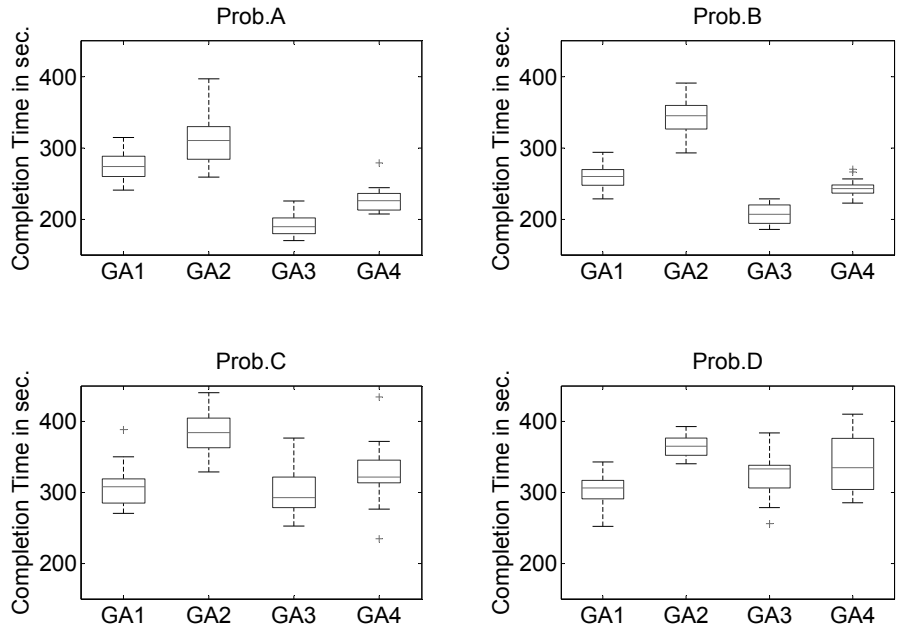}
\caption{The distribution of the solution quality of the subpopulation-based genetic algorithms with a single mutation operator (20 runs)}
\label{fig:2}
\end{figure}

This case study tests the performance of the subpopulation-based genetic algorithms with a single mutation operator (GA1--GA4 in Table~\ref{table:case2}); each mutation operator produces $pop\_sub-eli\_cnt=8$ offspring in each subpopulation.
The results are shown in Fig.~\ref{fig:2}. 
An ANOVA test shows that: (1) inversion (GA3) performs significantly better than the other three mutation operators when solving Prob.A and Prob.B; (2) the differences of the solution quality are not statistically significant when using swap, inversion, and displacement to solve Prob.C and Prob.D.

\subsection{Case study 3: Subpopulation-based GA with multiple mutation operators}
\begin{figure}[h!]
\centering
\includegraphics[width=\columnwidth]{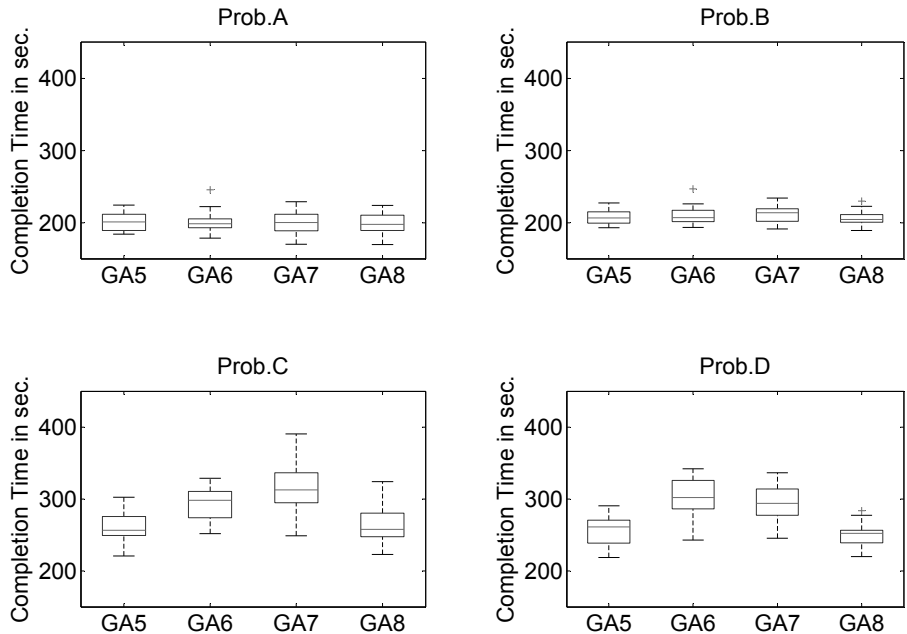}
\caption{The distribution of the solution quality of the subpopulation-based genetic algorithm with multiple mutation operators (20 runs)}
\label{fig:3}
\end{figure}

The third case study analyzes the performance of the subpopulation-based genetic algorithms with multiple mutation operators (GA5--GA8 in Table~\ref{table:case2}). 
Each mutation operator in GA5--GA7 produces $4$ offspring in each subpopulation by repeated application; each mutation operator in GA8 produces $2$ offspring in each subpopulation.
Inversion is combined with the other mutation operators in this case study, because it performed well in the second case study. The experimental results is displayed in Fig.~\ref{fig:3}. An ANOVA test shows that: (1) the differences of the solution quality between GA5--GA8 are not statistically significant when solving Prob.A and Prob.B; (2) GA5 and GA8 can provide significantly better solutions than GA6 and GA7 when solving Prob.C and Prob.D.

\begin{figure}[h!]
\centering
\includegraphics[width=\columnwidth]{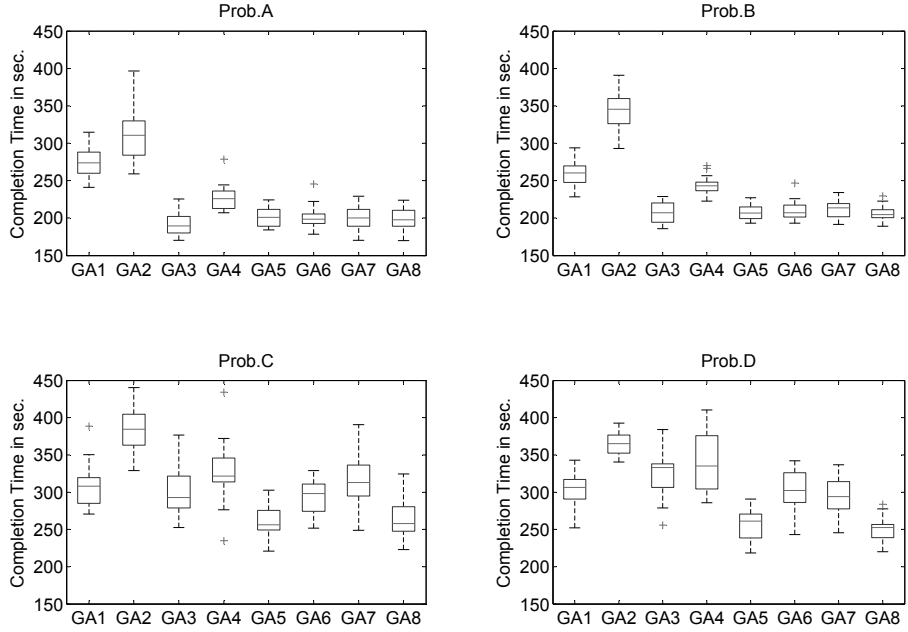}
\caption{The distribution of the solution quality of the subpopulation-based genetic algorithm with different mutation operators (20 runs)}
\label{fig:4}
\end{figure}

\begin{table}[h!]
\centering
\scriptsize{
\caption{Completion time $J$ in sec. for the subpopulation-based genetic algorithm with different mutation operators (best results highlighted in bold face)}
\label{table:caseall}
\begin{threeparttable}
\begin{tabularx}{\textwidth}{p{0.7cm}p{0.8cm}llllllll}
\hline
Problem & Criterion& GA1 & GA2 & GA3 & GA4 & GA5 & GA6 & GA7 & GA8  \\
\hline
Prob.A& $J_{\textsubscript{min}}$& 240.87&	259.00&	170.06&	207.28&	184.07&	178.41&	170.10&	\textbf{169.73}\\
& $J_{\textsubscript{mean}}$ & 273.98&	310.50&	\textbf{189.55}&	225.93&	200.86&	198.31&	200.00&	197.73\\
& $J_{\textsubscript{max}}$& 314.81&	396.67&	225.56&	278.89&	224.40&	245.39&	229.00&	\textbf{223.99}\\
\hline
Prob.B& $J_{\textsubscript{min}}$& 228.42&	293.00&	\textbf{185.95}&	222.65&	193.22&	193.25&	191.36&	189.00\\
& $J_{\textsubscript{mean}}$& 260.16&	345.23&	207.03&	243.05&	206.74&	206.91&	213.62&	\textbf{204.46}\\
& $J_{\textsubscript{max}}$& 294.05&	390.84&	228.75&	270.26&	\textbf{227.21}&	246.90&	234.23&	229.72\\
\hline
Prob.C&$J_{\textsubscript{min}}$ & 270.50&	328.82&	252.72&	234.96&	\textbf{220.76}&	251.82&	248.93&	222.82\\
&$J_{\textsubscript{mean}}$ & 308.14&	384.37&	292.78&	321.59&	\textbf{256.44}&	298.16&	312.96&	257.80\\
& $J_{\textsubscript{max}}$& 388.62&	440.42&	376.46&	434.31&	\textbf{302.74}&	328.96&	390.78&	324.31\\
\hline
Prob.D& $J_{\textsubscript{min}}$& 252.29&	340.35&	255.96&	285.73&	\textbf{218.58}&	242.96&	245.62&	220.00\\
& $J_{\textsubscript{mean}}$& 306.57&	365.23&	333.07&	335.02&	261.11&	302.09&	294.19&	\textbf{252.44}\\
& $J_{\textsubscript{max}}$&343.06&	392.85&	383.95&	410.17&	290.83&	342.09&	336.63&	\textbf{283.71}\\
\hline 
\end{tabularx}
\begin{tablenotes}
\item [1] GA1--Swap; GA2--Insertion; GA3--Inversion; GA4--Displacement; GA5--Swap and inversion; GA6--Insertion and inversion; GA7--Displacement and inversion; GA8--Swap, insertion, inversion, and displacement.
\item [2] Prob.A and Prob.B without cooperative tasks; Prob.C and Prob.D with cooperative tasks.
\end{tablenotes}
\end{threeparttable}}
\end{table}

The results of all tested subpopulation-based genetic algorithms listed in Table~\ref{table:case2} are shown in Fig.~\ref{fig:4} and Table~\ref{table:caseall}. GA3, GA5, and GA8 can provide better solutions than the other genetic algorithms. An ANOVA test shows that: (1) the differences of the solution quality using GA3, GA5, GA6, GA7, and GA8 are not statistically significant when solving Prob.A and Prob.B; (2) GA5 and GA8 perform significantly better than the other tested genetic algorithms when solving Prob.C and Prob.D. 

\subsection{Discussion}
In general, it is difficult to find the \textit{best} mutation operator that could produce all desired effects. The influences of mutation operators vary in difference genetic algorithms and in solving different problems. 
As illustrated above, inversion performs well to solve multi-robot task allocation problems without cooperative tasks, which is similar to the study of solving traveling salesman problems \citep{DeepM11, Albayrak2011, LiuKroll2012b}. The swap and inversion combination performs well to solve multi-robot task allocation problems with cooperative tasks. Swap and inversion are efficient in the simulation studies, which could be due to the fact that they can improve solution candidates with crossed paths effectively.

\begin{figure}[h!]
\centering
\includegraphics[width=0.8\columnwidth]{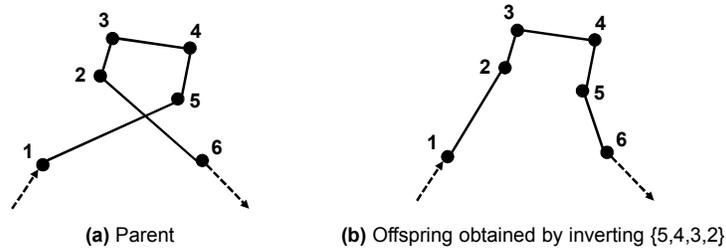}
\caption{An example with one cross for inversion}
\label{fig:iv}
\end{figure}

\begin{figure}[h!]
\centering
\includegraphics[width=0.8\columnwidth]{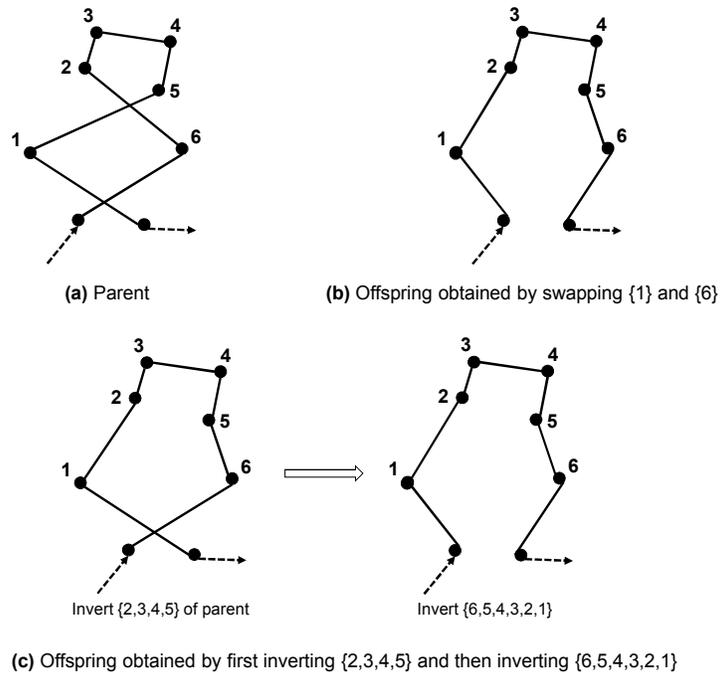}
\caption{An example with two crosses for swap and inversion}
\label{fig:sw}
\end{figure}

\begin{figure}[h!]
\centering
\includegraphics[width=0.8\columnwidth]{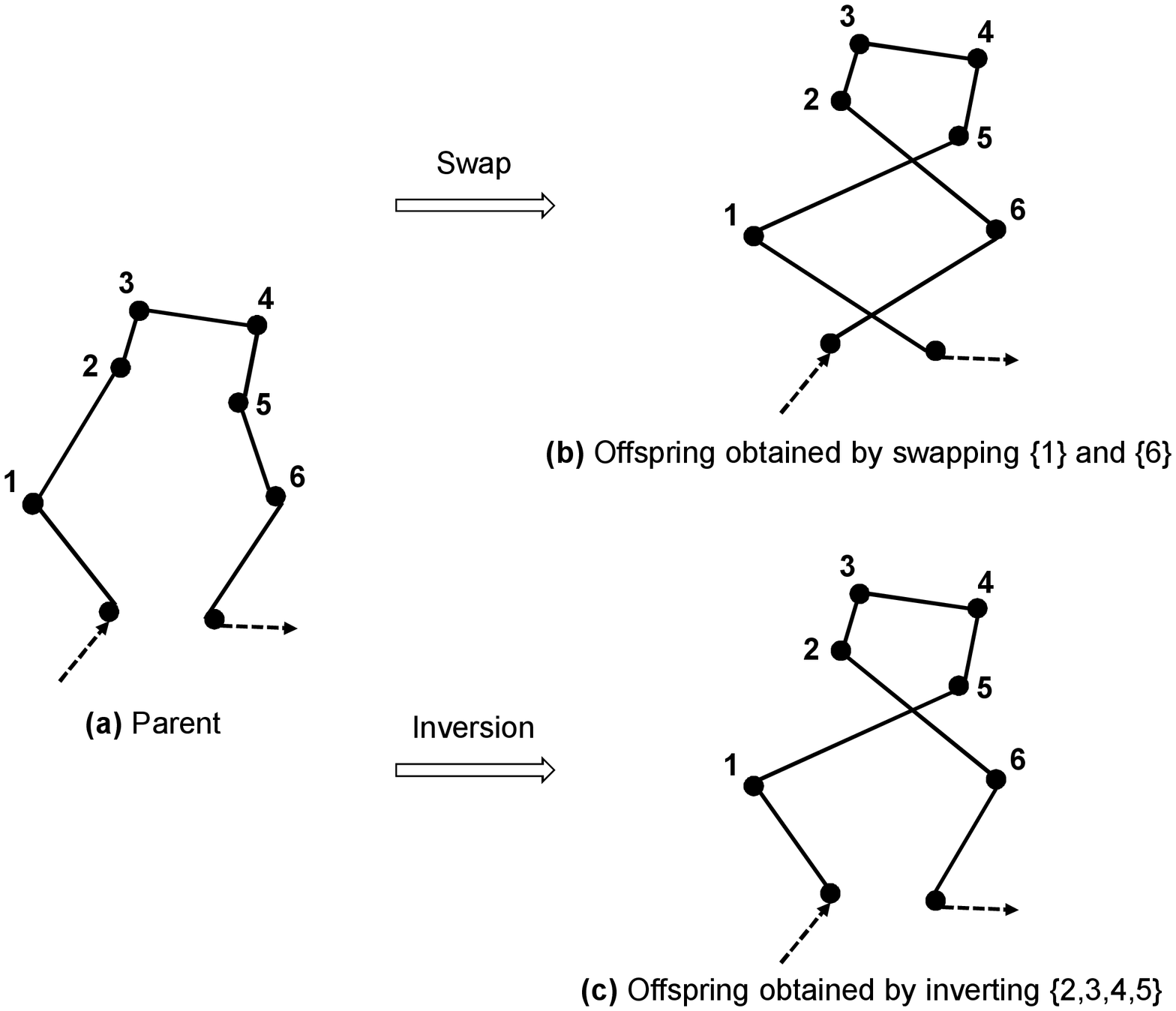}
\caption{An example of inappropriate swap and inversion}
\label{fig:swiv}
\end{figure}

In general, good solutions do not include crossed paths or only include a few crossed paths. 
Fig.~\ref{fig:iv}(a) shows an example where one cross may occur. This allocation can be improved by inverting $\{5,4,3,2\}$; see Fig.~\ref{fig:iv}(b). Fig.~\ref{fig:sw}(a) shows another example where two crosses may occur. This allocation can be improved by swapping $\{1\}$ and $\{6\}$; see Fig.~\ref{fig:sw}(b). 
If applying inversion to this allocation, the inversion mutation has to be used two times appropriately; see Fig.~\ref{fig:sw}(c).
These two examples imply that proper swap is more efficient than inversion in case of many crossed paths.
Insertion and displacement could not effectively improve these allocations.
On the contrary, inappropriate swap produces worse solutions than inversion, e.g. swap produces two crosses, while inversion produces one cross in Fig.~\ref{fig:swiv}.
Therefore, inversion can obtain better results than swap if given a large number of generations.

The mentioned improvement is observable from Fig.~\ref{fig:iv} and Fig.~\ref{fig:sw} when solving problems where the cost of one robot does not influence the costs of the other robots such as multi-robot task allocation problems without cooperative tasks. It becomes more complex in case of problems with cooperative tasks as the cost of finishing cooperative tasks does not depend on the task allocation for one robot but for two robots. Swap and inversion in Fig.~\ref{fig:iv} and Fig.~\ref{fig:sw} may cause a longer waiting time and a longer completion time.

\begin{figure}[h!]
\centering
\includegraphics[width=0.9\columnwidth]{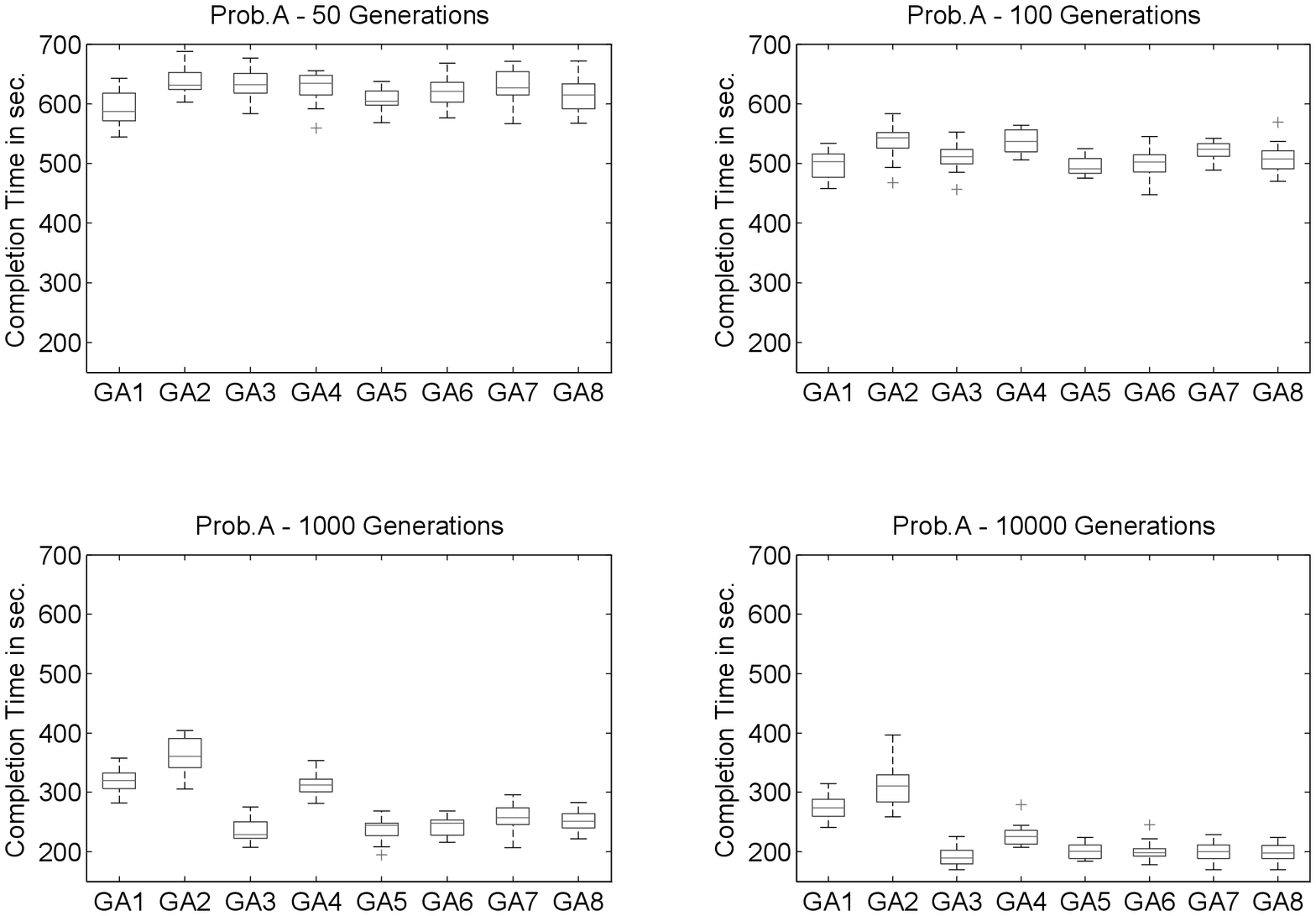}
\caption{The distribution of the solution quality of eight subpopulation-based genetic algorithms in different generations for solving Prob.A (20 runs)}
\label{fig:gennuma}
\end{figure}

\begin{figure}[h!]
\centering
\includegraphics[width=0.9\columnwidth]{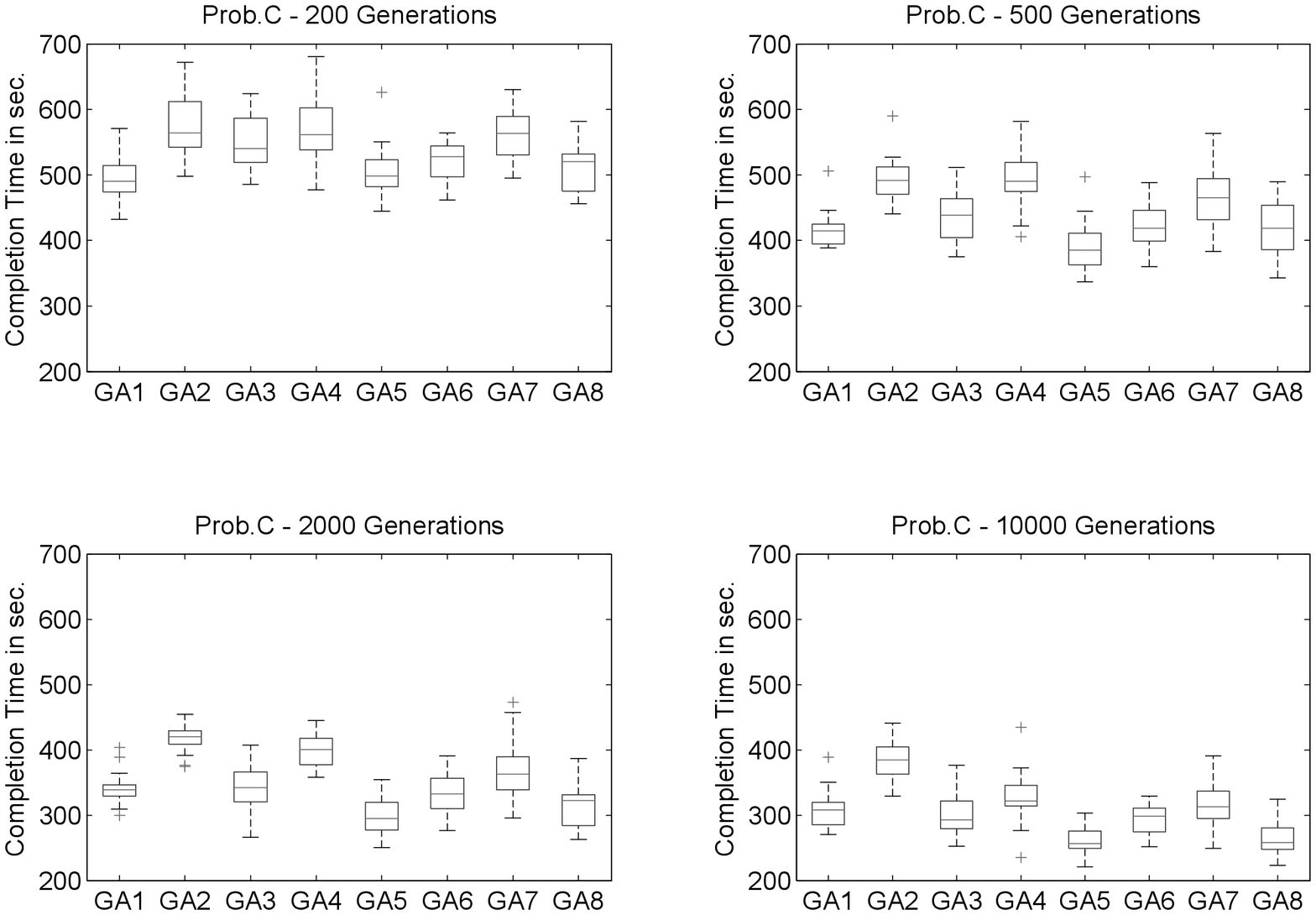}
\caption{The distribution of the solution quality of eight subpopulation-based genetic algorithms in different generations for solving Prob.C (20 runs)}
\label{fig:conv}
\end{figure}

To statistically evaluate the effect of different generations $gen\_num$, the distribution (20 runs) of the solution quality of eight subpopulation-based genetic algorithms in different generations is analyzed. The experimental results of solving problems without cooperative tasks (Prob.A and Prob.B) are similar: swap (GA1) obtains better solutions than inversion (GA3) within 100 generations, while inversion produces better solutions than swap after 100 generations; see Fig.~\ref{fig:gennuma}. 
In this part, we focus on the performance of genetic algorithms when solving problems with cooperative tasks (Prob.C and Prob.D). Similar results can be obtained when solving these two problems, e.g. Fig.~\ref{fig:conv}. 
The results indicate that swap (GA1) obtains better solutions than inversion (GA3) within 500 generations, while the swap-inversion combination (GA5) produces better solutions than swap after 500 generations. 
Many crossed paths may occur in the early generations due to the randomly generated initial population and a small number of generations. Hence, swap is more efficient than inversion in the early generations. 
%It practically demonstrates the above interpretation.
After 500 generations, the differences of the solution quality using swap and inversion are not statistically significant. The swap-inversion combination performs well because swap and inversion are applied to the same parent in each subpopulation. In this case, the good capabilities of both mutation operators are preserved.
As discussed before, multiple mutation operators involving swap and inversion are suggested to solve multi-robot task allocation problems, especially with cooperative tasks.

\section{Conclusion and outlook}
\label{sec:CON}
The problem complexity significantly increases if cooperative tasks are involved because they introduce additional spatial and temporal constraints.
In this paper, the performance of different mutation operators in a subpopulation-based genetic algorithm is analyzed for solving multi-robot task allocation problems without/with cooperative tasks. 
So far, a little work addresses this problem area.
The proposed subpopulation-based genetic algorithm uses just inversion mutation and selection, though obtains better solutions than classical genetic algorithms with tournament selection, partially mapped crossover (PMX), and inversion mutation in the test cases. Succeeding, a subpopulation-based genetic algorithm with four alternative mutation operators or with four mutation operator combinations was tested to find suitable mutation operators for multi-robot task allocation problems. The results indicate that inversion mutation performs well when solving problems without cooperative tasks, and a swap-inversion combination performs well when solving problems with cooperative tasks. 
As it is difficult to produce all desired effects with a single mutation operator, using multiple mutation operators is suggested, especially to solve complex problems.

The rate of each mutation operator is constant in this paper. As the performance of different mutation operators varies in different generations, future work will focus on employing an adaptive rate of mutation operators to improve the performance of the genetic algorithm.
Future work will also include more test scenarios, especially problems with a larger number of tasks.
The problem complexity increases with the number of robots required for each cooperative task.
In this paper, each cooperative task requires only two robots to carry it out simultaneously. In future work, the performance of the genetic algorithm will be analyzed when solving problems with cooperative tasks that require more than two robots to execute cooperatively.

\section*{Acknowledgements}
This work was supported by the scholarship awarded by the China Scholarship Council (CSC) and the Completion Scholarship awarded by the University of Kassel (Abschlussstipendien f\"{u}r Promovierende der Universit\"{a}t Kassel), which are greatly acknowledged.

\section*{Conflict of interest}
The authors declare that they have no conflict of interest.

\section*{References}

\bibliography{liubibfile}
%}
\end{document}